\documentclass{article}

\usepackage{ijcai26}

\usepackage{times}
\usepackage{soul}
\usepackage{url}
\usepackage[hidelinks]{hyperref}
\usepackage[utf8]{inputenc}
\usepackage[small]{caption}
\usepackage{graphicx}
\usepackage{amsmath}
\usepackage{amsthm}
\usepackage{booktabs}
\usepackage{algorithm}
\usepackage{algorithmic}
\usepackage[switch]{lineno}
\usepackage{amssymb}

\title{Skill-Latent Rewards: LLM-Guided Reward Formulation for Hierarchical
Reinforcement Learning}

\author{
Maryam Shoaeinaeini$^1$
\and
Brent Harrison$^1$\\
\affiliations
$^1$Department of Computer Science\\
University of Kentucky\\
Lexington, KY, USA\\
\emails
\{maryam.shoaei, brent.harrison\}@uky.edu
}

\begin{document}

\maketitle

\begin{abstract}
Long-horizon reinforcement learning under sparse and delayed feedback 
remains an open challenge. Existing return-decomposition methods 
redistribute episodic rewards to ease credit assignment, but they 
operate on raw state--action transitions and disregard the subgoal 
structure that governs meaningful progress in complex tasks. We 
propose \textsc{Skill-LR}, a hierarchical reinforcement learning 
framework in which a vision--language model identifies the agent's 
active subgoal at each timestep and a large language model generates 
a matching latent reward formulation through a two-stage prompting 
process. By conditioning reward signals on the currently active 
subgoal, \textsc{Skill-LR} produces stage-specific credit that 
evolves with the hierarchical plan, reducing temporal variance and 
improving sample efficiency. On multi-stage MiniGrid and BabyAI 
benchmarks, including DoorKey and Unlock-to-Unlock, 
\textsc{Skill-LR} converges faster and more reliably than sparse 
environmental rewards and strong credit-assignment baselines such as 
RUDDER and LaRe, demonstrating that subgoal-aligned latent rewards 
offer a principled path toward stable, interpretable credit 
assignment in long-horizon tasks.
\end{abstract}

\section{Introduction}

Despite the progress of reinforcement learning (RL) in continuous control and decision-making domains \cite{abdellatif2023reinforcement,kiran2021deep}, solving long-horizon tasks with sparse or delayed feedback remains an ongoing challenge \cite{hung2019optimizing,wang2025explicit}. In many environments, rewards are infrequent or available only at the end of an episode, making it difficult to determine which intermediate actions should be reinforced or suppressed \cite{luoma2017time,rahmandad2009effects}. This difficulty is fundamentally tied to the temporal credit-assignment problem, where the goal is to identify which actions meaningfully contributed to a delayed outcome \cite{zhang2020learning,sutton2011horde}. Prior research has attempted to mitigate these issues by relocating reward information closer to influential decisions within the trajectory. Return-decomposition methods such as RUDDER redistribute episodic rewards to specific state-action pairs \cite{arjona2019rudder}, while reverse RL \cite{zhang2020learning} and general value functions \cite{sutton2011horde} infer retrospective or predictive dependencies to improve temporal propagation. Recently, large language models (LLMs) have emerged as powerful tools for expressing high-level semantic intent, decomposing tasks into meaningful steps, and providing multifaceted evaluations. Building on this capability, recent latent-reward formulations \cite{qu2025latent} use LLMs to generate semantically enriched encodings of state-action pairs, which serve as structured inputs for proxy reward estimation and enable richer assessments than those derived from raw transition features. Although these approaches aid credit assignment in reinforcement learning, they operate primarily at the level of full episodes and, thus, fail to capture the fine-grained, stepwise progression required in complex tasks. 

In addition, long-horizon problems are composed of hierarchical subgoals. These structured dependencies imply that an action’s contribution must be evaluated according to how well it advances the current subgoal, rather than solely its influence on the final outcome. When reward signals do not reflect this hierarchical organization, actions that are critical for subgoal progress may receive little or no feedback, while irrelevant behaviors may be overemphasized simply because they precede the successful conclusion of a trajectory. Typical examples include navigating to a key before unlocking a door, assembling intermediate parts before completing a puzzle, or traversing regions in a required order before reaching a goal. In such settings, evaluating actions only by whether they achieve the overall goal obscures the subgoal-level structure that governs meaningful progress.

Hierarchical reinforcement learning (HRL) addresses environment complexity by decomposing long-horizon tasks into smaller, interdependent subproblems \cite{jiang2019language,pateria2021hierarchical}. By operating at multiple temporal scales, HRL reduces the effective horizon and promotes more structured exploration. However, early HRL methods relied on hand-crafted intrinsic rewards to train subpolicies. Examples of this include binary goal-achievement signals \cite{kulkarni2016hierarchical}, directional movement rewards \cite{vezhnevets2017feudal}, or state-goal distance functions \cite{nachum2018data}. Such predefined subtask rewards provide only coarse guidance and do not provide proportionate subgoal-aligned credit assignment, limiting their effectiveness in complex long-horizon environments. For HRL to be effective, reward mechanisms must accurately reflect subgoal-level progress and allocate credit in a manner consistent with the temporal dependencies that define hierarchical tasks.

Based on these observations, we pose the question core to the work presented in this paper:
Can we integrate the semantic interpretability of latent rewards with the structural alignment of hierarchical RL to achieve stable, temporally consistent credit assignment for long-horizon tasks?

To answer this, we introduce \textsc{Skill-LR}, a hierarchical RL framework that addresses the existing gap by coupling LLM-guided latent reward reasoning with skill-conditioned task decomposition. Rather than treating latent rewards as globally defined proxy values, \textsc{Skill-LR} conditions the latent reward computation on the agent’s current subgoal, enabling credit signals that evolve synchronously with the hierarchical plan in a long-horizon complex task. This design ensures that rewards assigned to intermediate actions reflect their contribution to local subgoal completion, ultimately supporting more effective progress toward the overall task. In essence, \textsc{Skill-LR} bridges semantic skill abstraction with temporally aligned reward redistribution, yielding a credit assignment strategy that is both interpretable and structurally consistent. To obtain a reliable LLM-based latent reward formulation, we introduce a two-stage prompting framework. In the first stage, the model performs iterative reasoning over the environment and its defined subgoals, generating and verifying multiple candidate reward formulations to ensure adaptability to the environment’s dynamics. In the second stage, these candidate formulations are re-evaluated through an optimization prompt, where the LLM compares prior outputs and selects the most coherent and hierarchically consistent latent reward structure. This process ensures that the final reward formulation aligns both semantically and temporally with the active subgoal hierarchy.

Based on the above inspiration, our contributions are summarized as follows:
\begin{itemize}
    \item We propose a skill abstraction predictor that leverages a vision–language model (VLM) to specify subgoal descriptions dynamically from the visual state at each timestep.
    \item We devise a hierarchical latent reward structure(\textsc{Skill-LR}) that conditions LLM latent credit formulation on subgoal transitions, ensuring temporally consistent and structurally coherent reward signals.
    \item We highlight the effectiveness of the \textsc{Skill-LR} model through a series of evaluations, demonstrating improvements in sample efficiency and convergence on multi-stage MiniGrid and BabyAI tasks.
\end{itemize}

Our evaluations show that \textsc{Skill-LR} also outperforms credit assignment based on ground-truth environmental rewards, highlighting the importance of integrating hierarchical reasoning with language-driven reward modeling. These findings suggest that aligning latent rewards with skill hierarchies is a promising direction for addressing the challenges of long-horizon RL.

\section{Related Work}
RUDDER \cite{arjona2019rudder} was introduced as a return-decomposition framework in which an LSTM-based return predictor identifies reward-causing transitions and redistributes terminal rewards into dense signals. This transformation improves sample efficiency in environments with delayed rewards. However, because RUDDER operates directly on raw state-action trajectories, its predictive model is limited by the low richness of the underlying inputs. Raw transitions may fail to capture the qualitative and quantitative factors that truly drive task success. To address this limitation, recent work \cite{qu2025latent} proposed the LaRe framework, which encode state-action information into higher-level semantic representations extracted via LLMs. Compared to raw features, these richer abstractions allow the reward model to reason over more meaningful environmental attributes, enabling stronger predictive structure. Despite their effectiveness, these approaches treat all transitions uniformly. They do not consider the structured ordering of subgoals that typically governs behavior in long-horizon tasks. These complex tasks are broken into smaller subtasks by HRL, allowing a high-level policy to solve the overall task by selecting which subtask to execute next \cite{pateria2021hierarchical,jiang2019language}. 

LLMs’ reasoning skills enable them to serve as effective planners, and they have shown strong performance in deriving meaningful skills \cite{huang2022language,ahn2022can}.  \cite{su2023subgoal} employs a vision-language model (VLM) to process state images and propose subgoals selected from a predefined library of skills. Similarly, \cite{shukla2023lgts} employs an LLM to generate subgoal sequences based on symbolic representations of the gridworld (e.g., positions and relations between objects), rather than raw visual inputs. 
However, in these cases, the role of the language model is restricted solely to subgoal selection. These methods neither incorporate LLM reasoning into latent reward redistribution nor use LLMs to derive subgoal-aware reward functions. As a result, they cannot provide semantically enriched or temporally aligned credit signals that reflect subgoal-level progress during training. These gaps motivate the development of our \textsc{Skill-LR} framework.

\section{Preliminaries}

\subsection{Markov Decision Processes}
RL problems are commonly formulated as Markov Decision Processes (MDPs), described by the tuple $\mathcal{M} = (\mathcal{S}, \mathcal{A}, P, r, \gamma)$, where $\mathcal{S}$ denotes the set of states, $\mathcal{A}$, the set of actions, $P$, the transition probabilities, $r$, the reward function, and $\gamma \in (0,1]$, the discount factor. At each timestep $t$, the agent observes a state $s_t \in \mathcal{S}$, takes an action $a_t \in \mathcal{A}$ following a policy $\pi(a_t \mid s_t)$, receives a reward $r_t$, and transitions to $s_{t+1}$. The objective is to find a policy $\pi(a_t \mid s_t)$ such that the expected discounted return or cumulative rewards, i.e., 
$R = \mathbb{E}_{\pi}\!\left[\sum_{t=0}^{T} \gamma^{t} r_t \right]$ be maximized.

In many environments, rewards are sparse and provided only at the end of an episode, making it difficult to determine which state-action pairs contributed to the final return. A central idea in return-decomposition methods such as RUDDER~\cite{arjona2019rudder} is to transform the original MDP into a \emph{return-equivalent} decision process in which the total return of a trajectory can be expressed as the sum of stepwise rewards \cite{zhang2023interpretable}, i.e., 
$R = \sum_{t=1}^{T} r(s_t,a_t)$.


In HRL, the standard MDP is augmented by introducing a space of subgoals or skills. HRL structure can be expressed as the tuple $(\mathcal{S}, \mathcal{G}, \mathcal{A}, P, r, \gamma)$, where $\mathcal{S}$, $\mathcal{A}$, $P$, $r$, and $\gamma$ retain their usual meanings, and $\mathcal{G}$ denotes the set of all possible subgoals. A high-level policy $\pi_H(g_t \mid s_t)$ selects a subgoal $g_t \in \mathcal{G}$, while a low-level policy $\pi_L(a_t \mid s_t, g_t)$ chooses primitive actions conditioned on both the current state and the assigned subgoal. The induced return under this hierarchical structure is given by $R = \mathbb{E}_{(g_t, a_t) \sim \rho_{\pi}} \left[ \sum_{t=0}^{T} \gamma^{t} \, r(s_t, a_t, g_t) \right]$,

and the objective is to find a pair of policies $(\pi_H, \pi_L)$ that maximizes this expected return.

\section{Methodology}

Humans acquire complex skills by decomposing tasks into manageable subtasks and using intermediate feedback to guide learning. 
In contrast, reinforcement learning agents typically optimize a single scalar reward, lacking explicit awareness of subgoal progress. 
To overcome this limitation, we propose \textsc{Skill-LR}, a framework that integrates language-driven subgoal abstraction with latent reward redistribution to capture both semantic and temporal structures in the reward space. 
As outlined in Algorithm~\ref{alg:skill-lr}, \textsc{Skill-LR} employs: 
(1) an LLM-based hierarchical planner to generate interpretable subgoal sequences, 
(2) a subgoal-conditioned latent reward formulation verified through environment interaction, and 
(3) a RUDDER-style return decomposition that redistributes latent rewards across trajectories, yielding temporally dense and semantically consistent signals for stable policy optimization.

\subsection{Theoretical Analysis}
\textbf{Proposition 1: Return Variance Reduction}

Assume the existence of a subgoal-conditioned latent transformation 
$\Phi: (s_t, a_t, g_t) \mapsto z_t^{g_t} \in \mathbb{R}^{D_g}$ 
that maps each state-action–subgoal tuple into a lower-dimensional latent subspace 
$\mathcal{Z}_{g_t}$, where the subspaces $\{\mathcal{Z}_{g_i}\}_{g_i \in \mathcal{G}}$ 
are approximately orthogonal, i.e., 
$\langle z^{g_i}, z^{g_j} \rangle \approx 0$ for all $i \neq j$.

Let $\hat{G}_t = f_{\textsc{LSTM}}\!\big(\Phi(s_t,a_t,g_t)\big)$ denote 
a subgoal-conditioned return predictor and define redistributed rewards as 
$\tilde{r}_t = \hat{G}_{t+1} - \hat{G}_t$.
The total episodic return can then be expressed as
\begin{equation}
    R = \sum_{t=1}^{T} \tilde{r}_t 
      = \sum_{g_i \in \mathcal{G}} R_{g_i},
\end{equation}
where each $R_{g_i}$ denotes the latent contribution of subgoal ~$g_i$ to the total episodic return.

Under the assumptions of 
(i) approximate temporal independence of redistributed rewards 
$\mathrm{Cov}[\tilde{r}_t, \tilde{r}_{t+k}] \approx 0$ for $k>0$, and 
(ii) orthogonal latent subspaces across subgoals 
$\mathrm{Cov}[R_{g_i}, R_{g_j}] \approx 0$ for $i \neq j$, 
the total return variance decomposes as
\begin{equation}
    \mathrm{Var}[R]
    = \sum_{g_i \in \mathcal{G}} \mathrm{Var}[R_{g_i}]
      + \sum_{g_i \neq g_j} \mathrm{Cov}[R_{g_i}, R_{g_j}]
    \;\approx\;
    \sum_{g_i \in \mathcal{G}} \mathrm{Var}[R_{g_i}].
\end{equation}
Because each $R_{g_i}$ is predicted from a lower-dimensional subspace 
$D_g \ll D$, the effective variance scales as
\begin{equation}
   \mathrm{Var}_{\textsc{Skill-LR}}[R]
   \;\approx\;
   \mathcal{O}(T D_g)
   \;\le\;
   \mathcal{O}(T D)
   \;=\;
   \mathrm{Var}_{LaRe}[R].
\end{equation}

\noindent
Compared to LaRe \cite{qu2025latent}, \textsc{Skill-LR} achieves reduced return variance through
latent subgoal decomposition, resulting in a tighter concentration bound, and thereby more stable gradient propagation.

\textbf{Proposition 2 (Latent-Space Regret Reduction).}
Let $\delta \in (0,1)$ and consider an RL agent trained over 
$K \in \mathbb{N}^+$ episodes in a hierarchical environment with horizon $T$.
In \textsc{LaRe}, the regret bound depends on the global latent dimension $D$ used 
to encode the full state--action space.  
Under \textsc{Skill-LR}, the hierarchical decomposition introduces a finite 
set of subgoals $\mathcal{G}$, where each subgoal 
$g \in \mathcal{G}$ induces a local latent subspace of dimension $D_g \ll D$.

Then, with probability at least $1 - \delta$, the cumulative regret of the 
hierarchical latent-space policy satisfies:
\begin{equation}
    \tilde{\mathcal{O}}\!\left(
        T\,|\mathcal{G}|\,\sqrt{D_g}\,
        \log\!\left(\frac{KT}{\delta}\right)
    \right)
    <
    \tilde{\mathcal{O}}\!\left(
        T\,D\,\sqrt{
            \log\!\left(\frac{KT}{\delta}\right)
        }
    \right)
\end{equation}
where $|\mathcal{G}|$ is the number of subgoals defining the hierarchical task,
$D$ is the global latent dimension used by \textsc{LaRe}, and $D_g$ is the 
subgoal-conditioned latent dimension introduced by \textsc{Skill-LR}. 

This inequality holds because hierarchical conditioning restricts the 
representation to a subgoal-specific manifold, reducing the effective latent dimensionality from $D$ to $D_g$, thereby tightening the regret bound and improving sample efficiency.

\subsection{Skill Abstraction and Prediction}
Manual subgoal specification is brittle and depends heavily on expert knowledge. To automate this process, we employ large language models (LLMs) guided by a structured prompt that defines the model’s role as a hierarchical planner, describes the environment, and specifies reasoning rules.

The LLM analyzes task dynamics and generates sequential subgoals (e.g., “pick up the key,” “open the door,” “reach the goal”) that are coherent and semantically grounded. Multiple decompositions are proposed and empirically evaluated-each used to train an agent, with the hierarchy yielding the highest average return selected as optimal.

Once the hierarchy is fixed, a vision-language model (VLM) identifies the active subgoal from visual observations. At each timestep $t$, it predicts:
\begin{equation}
\hat{g}_t = \arg\max_{g \in \mathcal{G}} f_{\textsc{VLM}}(s_t, g),
\end{equation}
where $f_{\textsc{VLM}}$ embeds the image state $s_t$ and subgoal text $g$ into a shared representation space and measures their compatibility. The predicted subgoal then conditions both the latent reward computation and hierarchical policy learning.

\begin{figure}[t]
    \centering
    \includegraphics[width=\columnwidth]{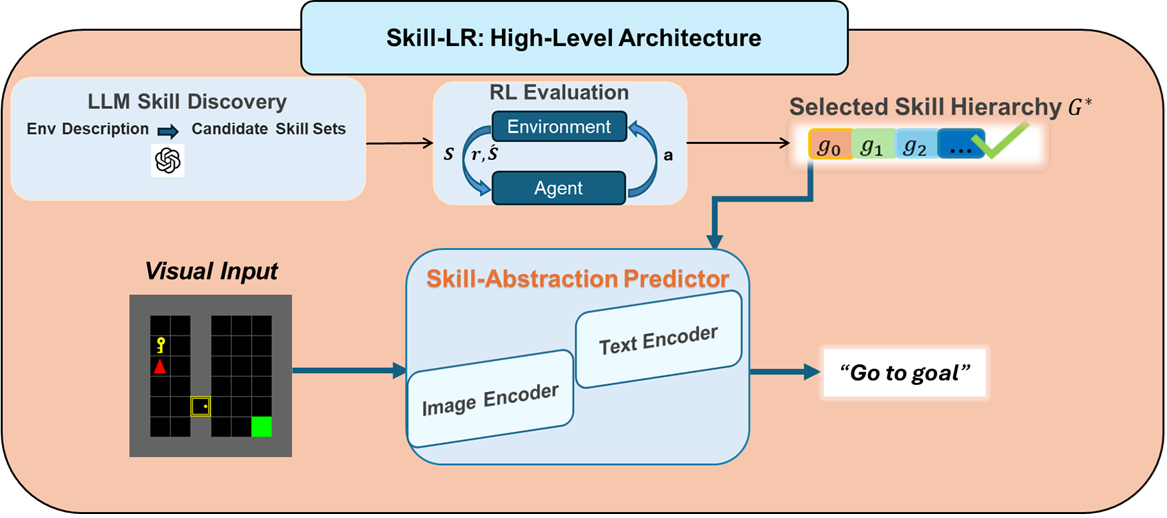}
    \caption{\textsc{Skill-LR} pipeline for skill discovery and abstraction.}
    \label{fig:Figure1}
\end{figure}
\begin{figure}[t]
    \centering
    \includegraphics[width=\columnwidth]{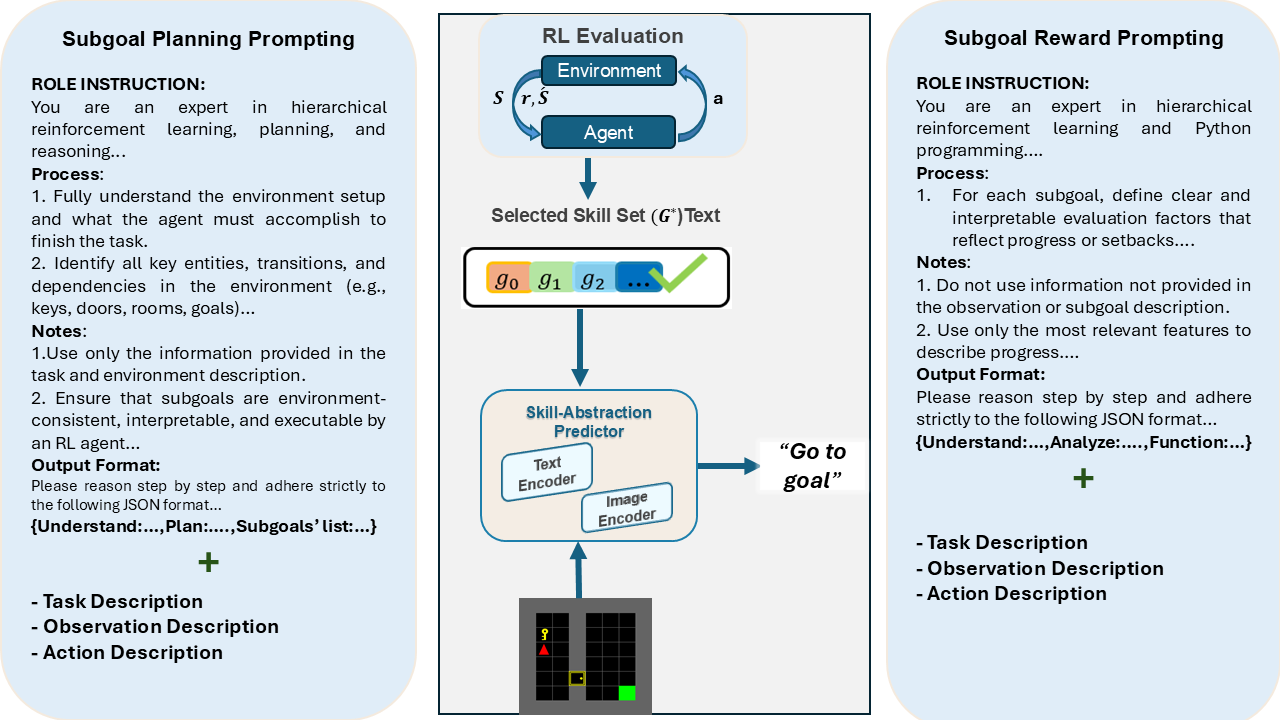}
    \caption{Two-stage prompting framework for latent reward formulation.}
    \label{fig:prompt_structure}

\end{figure}
Figure \ref{fig:Figure1} illustrates the high-level architecture of the proposed \textsc{Skill-LR} framework. The LLM-based Skill Discovery module generates candidate subgoal sets for DoorKey \cite{chevalier2018minimalistic} environment. The RL evaluation loop interacts with the environment to assess each candidate skill hierarchy, selecting the most consistent one. The Skill-Abstraction Predictor, composed of image and text encoders, integrates visual observations with subgoal representations to predict the active skill (e.g., “Go to goal”). 

\subsection{Skill-Conditioned Latent Reward Assignment}

LLMs can capture semantically rich environmental cues and produce informative evaluative signals \cite{qu2025latent}. 
To leverage this capability for hierarchical credit assignment, \textsc{Skill-LR} employs a structured prompting process that guides the LLM to reason about what should be rewarded or penalized in the context of subgoal progression. 
From multiple generated candidates, the model verifies and selects the reward function that best aligns with the intended subgoal semantics. 
This procedure, summarized in Algorithm~\ref{alg:skill-lr}, yields a high-quality latent reward specification conditioned on the current skill hierarchy.

\subsection{Reward Redistribution through Return Decomposition}

\textsc{Skill-LR} redistributes latent rewards rather than raw environmental signals. 
As  outlined in Algorithm\ref{alg:skill-lr}, After collecting complete subgoal-labeled trajectories $(s_t, a_t, g_t)$ stored in replay buffers for policy and reward updates, the LLM-based latent formulation evaluates each transition to produce 
semantic embeddings $\{z_t\}_{t=0}^{T}$.
An LSTM return predictor $F_\psi$ then estimates the cumulative latent return at each timestep, with redistributed rewards computed as:
\begin{equation}
\hat{r}_{t+1} = F_\psi(z_{t+1}) - F_\psi(z_t).
\end{equation}
This decomposition shifts sparse terminal rewards toward the intermediate decisions most responsible for subgoal completion, producing temporally dense, interpretable signals that accelerate convergence in long-horizon tasks.

\begin{algorithm}[tb]
\caption{\textsc{Skill-LR}: LLM-Guided Latent Reward and Return Redistribution}
\label{alg:skill-lr}
\textbf{Input:} LLM $\mathcal{M}$, environment $\mathcal{E}$, role instruction $\rho$, subgoal set $\mathcal{G}$, candidate number $n$, max episodes $N_{\max}$ \\
\textbf{Output:} Policy network $\pi_\theta$, latent reward predictor $F_\psi$
\begin{algorithmic}[1]
    \STATE Initialize policy parameters $\theta$, latent reward predictor parameters $\psi$, and buffers $\mathcal{B}_{\textsc{RL}}$, $\mathcal{B}_r$.
    \STATE Query $\mathcal{M}$ with $(\rho, \mathcal{E}, \mathcal{G})$ to generate $n$ candidate reward functions $\{R_i(s,a,g)\}_{i=1}^n$ aligned with subgoals.
    \STATE Evaluate each $R_i$ within $\mathcal{E}$ to verify functional correctness and environment consistency.
    \STATE Query $\mathcal{M}$ again with verified candidates to select the best reward function $R^*$ capturing subgoal progress semantics.
    \FOR{episode $= 1$ to $N_{\max}$}
        \STATE Collect trajectory $\tau = \{(s_t,a_t,g_t)\}_{t=0}^{T}$ under current policy $\pi_\theta$.
        \STATE Store $\tau$ in replay buffer $\mathcal{B}_{\textsc{RL}}$ and reward buffer $\mathcal{B}_r$.
        \IF{episode mod reward\_update\_frequency $= 0$}
            \WHILE{reward loss $> \textsc{threshold}$}
                \STATE Sample trajectory $\tau_i$ from $\mathcal{B}_r$.
                \STATE Compute latent embeddings: $\mbox{$z_t =\Phi(s_t,a_t,g_t; R^*)$}$.
                \STATE Train return predictor $F_\psi$ (RUDDER-style):
                \[
                    \hat{G}_t = F_\psi(z_t), \quad 
                    \hat{r}_{t+1} = \hat{G}_{t+1} - \hat{G}_t
                \]
                \STATE Update $F_\psi \leftarrow F_\psi - \alpha \nabla_\psi \mathcal{L}_{\textsc{RD}}(\psi)$.
            \ENDWHILE
        \ENDIF
        \STATE Update policy $\pi_\theta$ using redistributed latent rewards $\hat{r}_t$ from $\mathcal{B}_{\textsc{RL}}$.
        \STATE Reset replay buffer $\mathcal{B}_{\textsc{RL}}$.
    \ENDFOR
    \STATE \textbf{return} optimized policy $\pi_\theta$ and latent predictor $F_\psi$.
\end{algorithmic}
\end{algorithm}

\section{Experimental Setup and Results}

We evaluate \textsc{Skill-LR} on multi-stage MiniGrid \cite{chevalier2018minimalistic} and BabyAI \cite{chevalier2018babyai} benchmarks that require sequential reasoning and hierarchical decision-making. Our primary tasks include:

\begin{itemize}
    \item \textbf{DoorKey}: The agent must navigate to the key, pick it up, reach the locked door, open it, and proceed to the goal. This environment exhibits a clear hierarchical structure with visually grounded subgoals.
    \item \textbf{Unlock-to-Unlock}: A more challenging multi-stage variant in which the agent must unlock two doors in sequence. Specifically, the agent must first navigate to the initial key, use it to open the first door, enter the next room to obtain a second key, unlock the subsequent door, and finally locate the goal in the third room. This extended sequence of interdependent subgoals demands long-horizon temporal credit assignment and robust hierarchical reasoning.
\end{itemize}

Each environment provides pixel observations, which are processed by the skill-prediction module to identify subgoals at each timestep. Sparse rewards are given only upon episode completion, making the setting ideal for evaluating long-horizon credit assignment methods.

\begin{figure}[t]
    \centering
    \includegraphics[width=0.8\columnwidth]{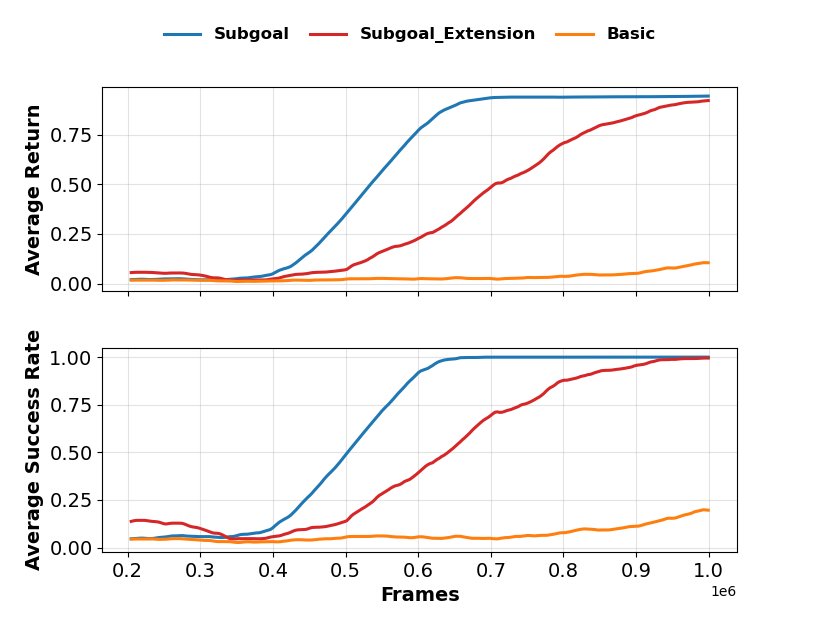}
    \caption{Candidate Skill Sets Evaluation in DoorKey environment}
    \label{fig:Figure2}
\end{figure}

All agents are trained using Proximal Policy Optimization(PPO) as the underlying RL algorithm. Experiments were conducted on an HPC node equipped with NVIDIA H200 GPUs, featuring 112 CPU cores and approximately 2 TB of RAM per node. In \textsc{Skill-LR}, skill abstraction is derived through the prompting pipeline described earlier. As shown in Figure~\ref{fig:Figure2}, RL evaluation on DoorKey identified a compact three-subgoal hierarchy
(\textit{pick up key} → \textit{open door} → \textit{reach goal})
as the most effective, achieving higher mean return and faster convergence than:  
(i) an extended five-subgoal decomposition (Subgoal\_Extension)(\textit{navigate to key} → \textit{pick up key} → \textit{navigate to door} → \textit{open door} → \textit{reach goal}),  
and (ii) basic fixed environment mission("go to goal") text as inputs to the RL agent.

For the \textsc{Unlock-to-Unlock} task, we adopt the sequence  
(\textit{pick up initial key} → \textit{open initial door} → \textit{pick up final key} → \textit{open final door} → \textit{go to goal}).  
To reduce task complexity, the \textit{drop} action is disabled in \textsc{Unlock-to-Unlock} task, allowing the agent to carry both keys simultaneously. All RL evaluations for choosing best hierarchy are implemented using the torch-ac RL starter codebase \cite{willems2019rlstarter}.
    
The \textsc{Skill-LR} setting for skill predictor and return decoder are as follows:
\begin{itemize}

\item A vision--language model (\textbf{CLIP ViT-L/14}) predicts
the active subgoal $g_t \in \{\texttt{PickupKey},\,\texttt{OpenDoor},\,
\texttt{GoToGoal}\}$ at each timestep by conditioning on the pixel
observation, grounding the hierarchical plan in the current visual
state. To adapt CLIP to the MiniGrid visual domain, we construct a
fine-tuning dataset by logging pixel observations together with the
environment's symbolic state and automatically deriving subgoal labels
via symbolic-state parsing. We parse the symbolic representation to
extract task-relevant predicates for the key, door, and goal (e.g.,
whether the agent is carrying the key, whether the door is open or
locked, and whether the goal has been reached), and deterministically
assign $g_t$ according to these predicate transitions: states are
labeled \texttt{PickupKey} until key possession becomes true,
\texttt{OpenDoor} until the door becomes open, and \texttt{GoToGoal}
until the agent reaches the goal. The symbolic state is used only to
generate training labels; the learned subgoal predictor operates
solely on pixels at inference time.

We fine-tune CLIP online with a lightweight setup by freezing the
pretrained backbone and updating only the image projection head, while
keeping the text embeddings for the three subgoals fixed. Training
uses replay-buffered image--subgoal pairs collected during interaction;
the model is updated every 10 episodes for 10 gradient steps with
minibatches of size 256. CLIP adapts quickly: after only 3 updates
(30 gradient steps), average training accuracy rises from 80.9\% to
100\% and loss drops from 0.527 to 0.014. Notably, the fine-tuned
model also shows stronger semantic subgoal recognition than the
symbolic label in edge cases. For example, after the door is opened
and the key is dropped in the first room, CLIP correctly identifies
the active subgoal as \texttt{GoToGoal}, whereas the symbolic state
still incorrectly indicates \texttt{PickupKey}.

\item For \textsc{Unlock-to-Unlock}, we follow the same
symbolic-state parsing procedure but extend the object set to reflect
the two-stage unlocking structure. We track \texttt{initial\_key},
\texttt{initial\_door}, \texttt{final\_key}, \texttt{final\_door},
and \texttt{goal}, and assign subgoal labels based on the
corresponding predicate transitions: collecting the relevant key,
opening the associated door, and ultimately reaching the goal. The
optimized skill set is retrieved from an LLM as described in
Section~\ref{sec:method}.

\item A \textbf{return-decomposition LSTM} converts
trajectory-level latent rewards into dense redistributed rewards via
return decomposition. We employ a single-layer LSTM (hidden
size~$= 64$) to encode the full trajectory sequence and output a
scalar prediction at each timestep via a linear head.

\end{itemize}

In DoorKey and Unlock-to-Unlock, we run all experiments over four and three random seeds, respectively, and report the mean episodic return and success rate, with 95\% confidence intervals across seeds. 
 
\subsection{Baselines}
We compare \textsc{Skill-LR} against several competitive credit-assignment baselines as follows:

\paragraph{PPO with Environmental Reward.}
A standard PPO agent trained solely on sparse environmental rewards. This baseline highlights the difficulty of long-horizon credit assignment without any auxiliary shaping or semantic feedback.

\paragraph{RUDDER} \cite{arjona2019rudder}.
A reward-redistribution method that decomposes episodic returns using an LSTM-based return predictor to assign dense contributions to individual timesteps. RUDDER operates directly on raw state-action trajectories, testing whether latent state-action transformation can boost credit assignment.

\paragraph{\textsc{LaRe} (Latent Reward)} \cite{qu2025latent}.
A recent latent-reward-based approach that employs LLM reasoning to extract semantic reward factors from raw state--action pairs. Unlike \textsc{Skill-LR}, LaRe does not condition its latent rewards on subgoals, providing a comparison to non-hierarchical latent reward formulations.

\paragraph{Global LLM Latent Reward.}
A naive non-hierarchical variant of \textsc{Skill-LR}, where the LLM evaluates each transition with respect to the entire task rather than the active subgoal predicted by the VLM. While this yields semantically informed latent rewards, the resulting signal lacks the subgoal-specific temporal alignment introduced by \textsc{Skill-LR}. This baseline isolates the contribution of subgoal conditioning in hierarchical latent reward computation.


\subsection{Results}

We evaluate \textsc{Skill-LR} and all baselines across hierarchical MiniGrid and BabyAI benchmarks to assess three key aspects: (i) sample efficiency under sparse reward settings, and (ii) stability of policy convergence across random seeds.

\begin{figure}[t]
    \centering
    \includegraphics[width=\columnwidth]{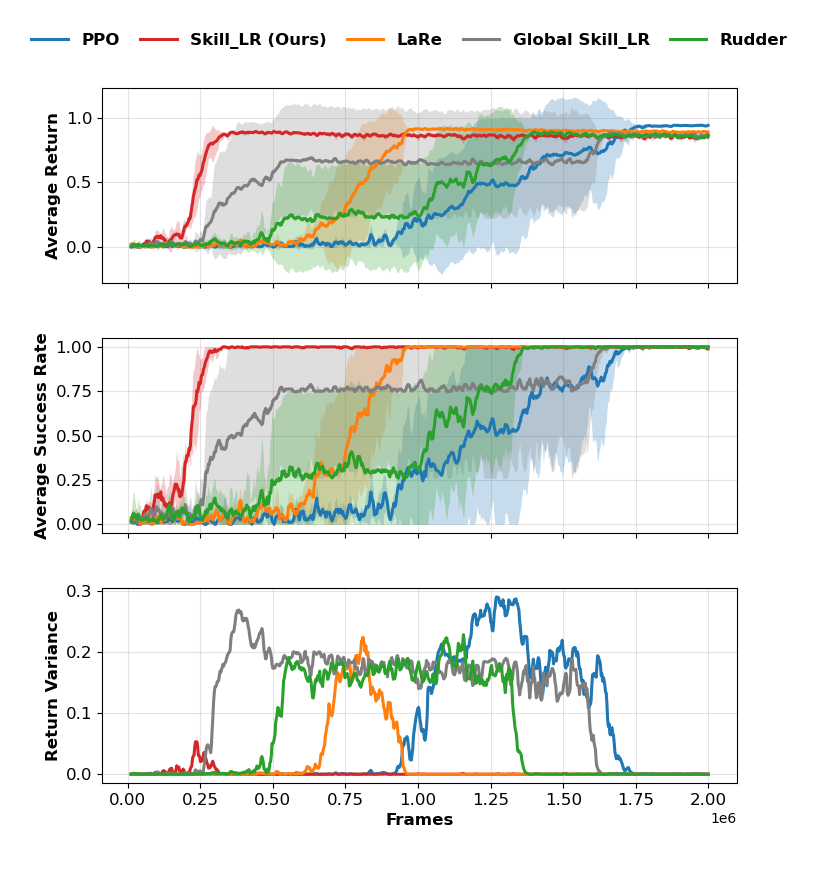}
   \caption{Performance and training-stability comparison on MiniGrid DoorKey 8$\times$8 for PPO, \textsc{Skill-LR} (ours), LaRe, Global Skill\_LR, and RUDDER. The top panel shows average episodic return, the middle panel shows average success rate, and the bottom panel shows the inter-seed variance of training return. All curves are computed over four random seeds (1, 10, 20, 30), with shaded regions indicating 95\% confidence intervals across seeds.}
    \label{fig:Figure3}
\end{figure}

Figure \ref{fig:Figure3} shows the learning curves of all methods on the 8$\times$8 DoorKey environment, averaged over four random seeds. We report both average episode return and success rate, which provide complementary views of learning quality and reliability. \textsc{Skill-LR} (red) achieves the fastest and most consistent improvement among all approaches. Within the first few hundred thousand frames, it reaches near-optimal return and success rate with a substantially narrower confidence band than the baselines, indicating more stable convergence across seeds.

In contrast, the global latent baseline (Global Skill\_LR, gray) exhibits slower improvement and noticeably higher variability throughout training. This suggests that using a single global latent reward over the full task introduces a less stable learning signal than conditioning rewards on the currently active subgoal. RUDDER (green) eventually achieves strong performance, but converges more slowly and with greater variability, highlighting the difficulty of redistributing sparse rewards without explicit hierarchical structure. LaRe (orange) also improves steadily, but remains slower and less stable than \textsc{Skill-LR}, indicating that a global latent formulation is less effective than subgoal-conditioned reward shaping in this multi-stage setting. PPO using only the environment reward (blue) shows the slowest progress and the widest confidence band, reflecting the challenge of learning from sparse terminal rewards alone.

The success-rate curves further emphasize this difference. \textsc{Skill-LR} reaches near-perfect success substantially earlier than all baselines, showing that subgoal-conditioned latent rewards provide more timely and informative credit signals during training. While several baselines eventually approach strong asymptotic performance, they require substantially more interaction frames, indicating lower sample efficiency.

To further assess training stability, we compute the inter-seed variance of training return at each frame across four random seeds. In sparse-reward tasks, this variance is naturally close to zero at the beginning of training because all seeds obtain similarly low return. As learning begins, the variance increases because different seeds improve at different times; once training converges, it decreases again as the learned policies become more similar. The variance plot in Figure \ref{fig:Figure3} shows that \textsc{Skill-LR} has a narrower and lower high-variance region than PPO, RUDDER, LaRe, and Global Skill\_LR, indicating faster and more consistent convergence across seeds. Taken together, these results provide empirical evidence that subgoal-conditioned latent reward redistribution improves training stability.

\begin{figure}[t]
    \centering
    \includegraphics[width=\columnwidth]{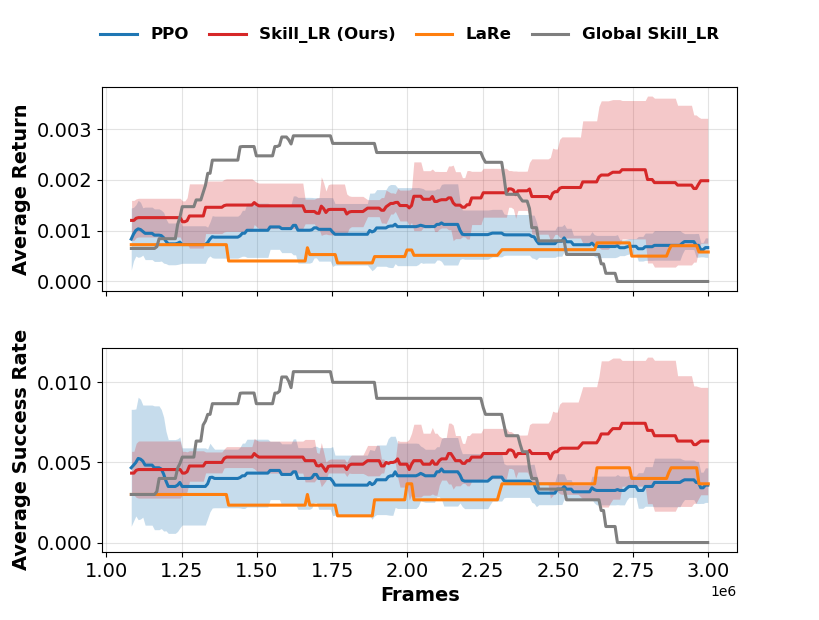}
    \caption{Learning curves on BabyAI Unlock-to-Unlock comparing PPO with episodic environmental reward, Skill\_LR (Ours), LaRe, Global Skill\_LR. Top: average episodic return. Bottom: average success rate. Curves show the mean over 3 random seeds (1, 10, 20), with shaded regions indicating 95\% confidence intervals across seeds.}
    \label{fig:Figure4}
\end{figure}

Figure \ref{fig:Figure4} presents the results on the Unlock-to-Unlock environment. \textsc{Skill-LR} and PPO were the only methods that trained successfully across all three seeds. In contrast, LaRe and Global Skill-LR completed training under only a single seed, as their LSTM-based return-prediction modules produced NaN values and diverged in the remaining runs. RUDDER was unable to train under any seed, as its return-decomposition LSTM became numerically unstable after only a few updates. It is therefore omitted from the Figure. \textsc{Skill-LR} achieves the highest and most stable performance in both average return and success rate, demonstrating that subgoal-conditioned latent rewards remain informative even in deep hierarchical settings. PPO exhibits gradual but much slower improvement, consistent with the difficulty of solving this multi-stage task using only sparse environmental rewards. LaRe and Global Skill-LR show weak or unstable learning dynamics.Because these methods apply a global latent reward function conditioned on all subgoals simultaneously, they fail to provide the temporally aligned subgoal-specific credit signals necessary for stable policy updates. The resulting gradients are noisy and often contradictory, explaining their collapse in two of the three seeds. 

\begin{figure}[t]
    \centering
    \includegraphics[width=\columnwidth]{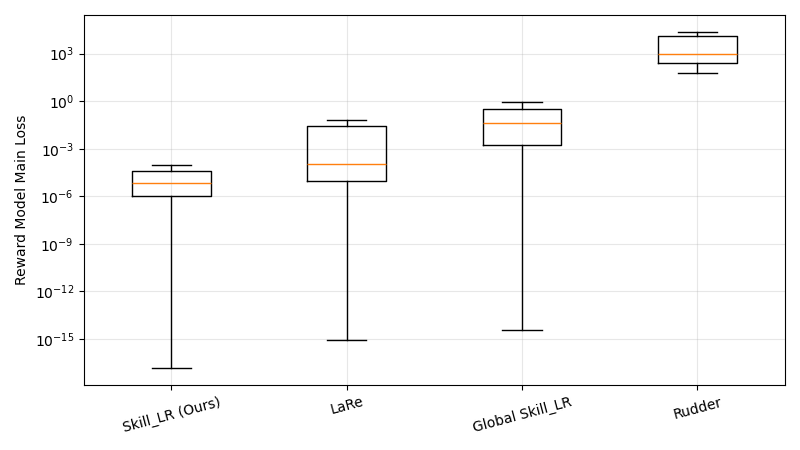}
    \caption{Distribution of reward-model main loss on the Unlock environment across training updates. Lower values indicate more accurate final-step return prediction. Skill-LR achieves the lowest loss distribution among all methods, followed by LaRe, while Global Skill\_LR and RUDDER exhibit substantially larger losses. This suggests that subgoal-conditioned latent reward decomposition yields a more stable and learnable return-modeling objective in more challenging hierarchical tasks.}
     \label{fig:quatile_unlock_main-loss}
\end{figure}

Figure \ref{fig:quatile_unlock_main-loss} shows the distribution of reward-model main loss on the Unlock environment. Unlike DoorKey, where the loss distributions are broadly similar across methods, Unlock reveals a much clearer separation. Skill-LR achieves the lowest overall main-loss distribution, followed by LaRe, while Global Skill-LR and RUDDER exhibit substantially larger losses. This indicates that, in a more challenging hierarchical setting, subgoal-conditioned reward decomposition yields a more accurate and stable return-modeling objective. These results are consistent with the downstream performance gains observed in return and success rate in Figure \ref{fig:Figure4}.

\section{Discussion}
\textbf{Does integrating semantic latent rewards with a hierarchical skills structure improve temporal credit assignment?} Our results provide a clear affirmative answer to the motivating question of this work. \textsc{Skill-LR} demonstrates that integrating LLM-derived state encodings with hierarchical task decomposition produces reward signals that are substantially more informative than those obtained from raw state-action transitions or latent encodings itself. By conditioning latent rewards on the agent’s active subgoal, \textsc{Skill-LR} allocates credit in a manner that is both semantically grounded and temporally aligned with the structure of the task. This alignment is essential in long-horizon environments like Unlock to Unlock environment, where delayed outcomes span multiple decision stages and conventional redistribution methods struggle to distinguish which transitions correspond to which stage of progress.

\textbf{Does \textsc{Skill-LR} improve sample efficiency compared to sparse environmental rewards?} These evaluations suggest that \textsc{Skill-LR} does improve sample efficiency. Our experiments show \textsc{Skill-LR} learns significantly faster than the environmental reward baseline, even though all methods eventually approach similar final returns in DoorKey. Its advantage comes from the early identification and reinforcement of subgoal-relevant transitions, which provide dense and timely learning signals throughout training. Although Figure~\ref{fig:Figure3} shows that the environmental reward baseline-or methods focused solely on the final mission-can eventually reach the maximum return and in some cases exceed \textsc{Skill-LR} by roughly 5\%, their slow early-stage improvement highlights a substantial sample-efficiency gap that \textsc{Skill-LR} decisively overcomes.

\section{Conclusion}
In this work, we introduced \textsc{Skill-LR}, a hierarchical reinforcement learning framework that integrates large language models into both skill abstraction and latent reward formulation. By combining LLM-derived state encodings with accurate prediction of the agent’s active subgoal, \textsc{Skill-LR} constructs semantically enriched representations that align naturally with the hierarchical structure of long-horizon tasks. This alignment enables more effective and temporally consistent credit assignment.

Our contributions in this paper are threefold: (i) we address an understudied challenge in credit assignment by enabling subgoal-aligned reward redistribution in environments where meaningful progress unfolds across multiple decision stages; (ii) we develop a practical algorithm that is both empirically and theoretically validated, demonstrating substantial improvements in sample efficiency, stability, and convergence across MiniGrid and BabyAI benchmarks; and (iii) we advance the role of LLM reasoning in reinforcement learning by unifying hierarchical task decomposition with semantically grounded reward modeling.

The experimental results show that conditioning latent rewards on active subgoals not only accelerates learning but also stabilizes policy updates in complex long-horizon settings. We believe that \textsc{Skill-LR} offers a promising foundation for future work at the intersection of hierarchical RL and language-informed structure. Potential extensions include applications to continuous control, robotic manipulation, and multi-agent coordination, where structured subgoal reasoning and semantically informed rewards could yield further gains.

\bibliographystyle{named}
\bibliography{ijcai26}

\end{document}